\pdfoutput=1
\documentclass[11pt]{article}
\usepackage{booktabs}
\usepackage{multicol}
\usepackage[final]{acl}
\usepackage{times}
\usepackage{latexsym}
\usepackage[T1]{fontenc}
\usepackage[utf8]{inputenc}
\usepackage{microtype}
\usepackage{inconsolata}
\usepackage{amsmath}
\usepackage{tcolorbox}

\usepackage{graphicx}

\usepackage{todonotes}
\title{Decoding Emergent Big Five Traits in Large Language Models: Temperature-Dependent Expression and Architectural Clustering}

\author{
Christos-Nikolaos Zacharopoulos\thanks{\ \ Corresponding author.} \\
Independent Researcher \\
\texttt{christonik@gmail.com} \\
\And
Revekka Kyriakoglou \\
Université Paris 8 \\ Vincennes -- Saint-Denis, Paris, France \\
\texttt{revekka.kyriakoglou@univ-paris8.fr} \\
}

\begin{document}
\maketitle
\begin{abstract}
As Large Language Models (LLMs) become integral to human-centered applications, understanding their personality-like behaviors is increasingly important for responsible development and deployment. This paper systematically evaluates six LLMs, applying the Big Five Inventory-2 (BFI-2) framework, to assess trait expressions under varying sampling temperatures. We find significant differences across four of the five personality dimensions, with Neuroticism and Extraversion susceptible to temperature adjustments. Further, hierarchical clustering reveals distinct model clusters, suggesting that architectural features may predispose certain models toward stable trait profiles. Taken together, these results offer new insights into the emergence of personality-like patterns in LLMs and provide a new perspective on model tuning, selection, and the ethical governance of AI systems. We share the data and code for this analysis here: \url{https://osf.io/bsvzc/?view_only=6672219bede24b4e875097426dc3fac1}
\end{abstract}

\section{Introduction}
The increasing use of Large Language Models (LLMs) as substitutes for human interaction marks a significant shift in societal dynamics. Individuals now engage with LLMs not only for retrieving information but also in contexts resembling intimate human dialogue, including seeking emotional support, personal advice, and even therapeutic guidance \cite{stade2024large, li2023systematic}. As these disembodied interactions become more commonplace, LLMs are beginning to occupy roles once reserved for human experts, counselors, or friends. This development has profound implications: by altering traditional models of communication, mental health care, and interpersonal relationships, LLMs challenge established norms of trust, empathy, and reliability. Moreover, as these systems become more human-like in their conversational styles, users increasingly anthropomorphize them, projecting cognitive and emotional qualities onto what are, at their core, statistical models. Such anthropomorphization raises critical questions about the nature of “personality” in LLMs and how users may rely on these perceived traits when forming judgments, seeking comfort, or making important decisions.

In response to these emergent issues, researchers have begun probing whether and how LLMs exhibit human-like personality characteristics \cite{serapio2023personality, jiang2023personallm, mao2023editing, zhan2024humanity, noh2024llms}. Anchoring such inquiries in robust psychological frameworks helps clarify otherwise nebulous concepts. The Big Five personality model—capturing openness, conscientiousness, extraversion, agreeableness, and neuroticism \cite{mccrae1997}—serves as a widely accepted and empirically supported tool for understanding human personality. Although its application to LLMs is still in its infancy, a growing body of work suggests that LLMs may indeed reflect trait-like patterns in their generated responses \cite{lee2024llms}. Understanding these patterns is far from a purely academic exercise; it has far-reaching implications for the design, deployment, and ethical governance of AI-driven communication.

Despite initial efforts, critical gaps remain. Existing literature has primarily focused on whether traits like those in the Big Five emerge in LLMs, but not on the underlying mechanisms that give rise to these traits or the conditions that influence their stability. For instance, there is limited insight into how model architecture, training data composition, and sampling strategies interact to shape the personality-like behaviors observed. Within the broader research effort to contextualize the nature of LLM “personality,” examining additional factors, like the temperature parameter, can provide fresh perspectives.

This paper contributes new analytical depth along two axes. (1) We systematically examine trait expression under varying sampling temperatures to characterize how a core decoding control modulates personality-like outputs. (2) We use agglomerative hierarchical clustering to uncover model-level patterns of similarity in trait profiles, providing evidence of structural tendencies across architectures. Together, these analyses move beyond simple personality testing and clarify how model design and decoding interact to shape personality-like behaviors.

We advance this exploration by evaluating six comparably sized LLMs using the Big Five Inventory–2 (BFI-2) questionnaire \cite{soto2017next}, a validated and reliable measure of human personality traits. Beyond simply classifying the presence or absence of trait-like patterns, we systematically vary the temperature parameter to investigate its role as a stochastic decoding control that may modulate responses. We also attempt to identify if and how LLMs cluster natively as a factor of their personality responses. Through this multifaceted analysis, we aim to deepen our understanding of what it means for LLMs to exhibit personality-like traits, identify the factors that modulate these expressions, and lay the groundwork for more accountable and human-centered design and governance of AI communication systems.

\section{Methods}

We employed a diverse ensemble of state-of-the-art large language models (LLMs), each fine-tuned for conversational tasks and equipped with distinct attention mechanisms. The selected models spanned varying parameter scales, vocabulary sizes, and attention mechanisms. We utilized the \textit{Llama 3 8B} model from the Llama series, featuring 8 billion parameters, a vocabulary size of 128,256 tokens, and Grouped-Query Attention (GQA) \cite{llama3}. The \textit{Mistral 7B} model, with 7.3 billion parameters and a vocabulary size of approximately 131,000 tokens, incorporated Grouped-Query Attention (GQA) \cite{mistral}. The \textit{MythoMax L2 13B} model from the Gryphe series combined 13 billion parameters with an 8,000-token context length \cite{mythomax}. The \textit{Gemma 9B} model, with 9 billion parameters and a vocabulary size of 300,000 tokens, employed dynamic attention scaling \cite{gemma}. The Q\textit{wen 7B} model utilized a vocabulary of over 150,000 tokens alongside sliding window attention \cite{qwen}. Lastly, the \textit{StripedHyena 7B} model, featuring 7 billion parameters and a vocabulary size of 280,000 tokens, implemented block-sparse attention \cite{stripedhyena}. To systematically evaluate the influence of sampling temperature on model outputs, we conducted a series of text-generation experiments using a fixed prompt and instructions designed to simulate a personality test response scenario. Specifically, we varied the temperature parameter from 0 to 2 in increments of 1, yielding a total of 21 experimental conditions for each of the 60 main questions of the BFI-2 questionnaire. This range was chosen to capture an extensive spectrum of possible sampling behaviors, from highly deterministic (temperature = 0) to increasingly stochastic regimes. Nevertheless, this range falls within a reasonable space of exploration. %To constrain verbosity and maintain concise outputs, we limited the maximum number of tokens per response to 20 (\texttt{max\_tokens=20}). To balance diversity and coherence, we set \texttt{top\_k=60}, ensuring that at each generation step, only the 60 most probable tokens were considered, and employed a nucleus sampling threshold of \texttt{top\_p=0.8} to dynamically filter candidates based on cumulative probability. This combination enabled exploring a broad range of token candidates without drifting into low-probability, incoherent outputs. We applied a repetition penalty of 1.1 to discourage repetitive or trivial patterns, incrementally penalizing repeated phrases and promoting more varied surface forms. Finally, all responses were truncated upon encountering either the \texttt{"<|eot\_id|>"} or \texttt{"<|eom\_id|>"} special tokens, ensuring that the generated responses remained focused on providing a single numerical value reflecting the subject’s degree of agreement or disagreement with the given statement.

Statistical analysis comprised three main components. First, we performed non-parametric between-model across personality type comparisons using Kruskal-Wallis H-tests ($\alpha = 0.05$) to detect significant differences in trait expressions. %followed by post-hoc pairwise Mann-Whitney U tests with Bonferroni correction ($p_{\text{adjusted}} = p \cdot m$, where $m$ is the number of comparisons) and rank-biserial correlation coefficients ($r_{\text{rb}}$) for effect size quantification. 

Second, we conducted temperature sensitivity analyses through multiple linear regression models for each trait dimension, with temperature as the predictor variable and trait scores as the response variable, complemented by Pearson correlation coefficients ($r$) to assess relationship strength and direction.
The regression analyses aim to quantify the associations between temperature and trait scores, rather than implying causal relationships or theoretical psychological mappings.

Third, we employed agglomerative hierarchical clustering with Ward's minimum variance method
%$
%d = \sqrt{\sum (x_i - y_i)^2}
%$
using Euclidean distance metrics to identify model groupings. These groupings were validated through trait covariance matrices
%$
%\Sigma_{ij} = E\left[(X_i - \mu_i)(X_j - \mu_j)\right]
%$
to examine inter-trait relationships.

\begin{table}[ht]
\centering
\begin{tabular}{lcc}
\toprule
\textbf{Domain}               & \textbf{Statistic} & \textbf{p-value}  \\
\midrule
Extraversion          & 40.7803 & $<$0.01 \\
Agreeableness         & 65.3067 & $<$0.01 \\
Conscientiousness     & 63.0415 & $<$0.01 \\
Neuroticism           &  9.2691 & n.s.   \\
Openness to Experience & 58.1957 & $<$0.01 \\
\bottomrule
\end{tabular}
\caption{Kruskal-Wallis test results for personality domains. The table shows the test statistic and corresponding p-values for each domain.}
\label{table1}
\end{table}

\begin{table}[ht]
\centering
\small
\begin{tabular}{lccc}
\toprule
\textbf{Domain} & \boldmath{$R^2$} & \textbf{Pearson Cor.} & \textbf{p-value} \\
\midrule
Neuroticism & 0.3486 & -0.5904 & $<$0.05 \\
Extraversion & 0.2521 & 0.5021 & $<$0.05 \\
Agreeableness & 0.0343 & -0.1853 & n.s. \\
Conscientiousness & 0.0257 & 0.1602 & n.s. \\
Openness & 0.0003 & 0.0178 & n.s. \\
\bottomrule
\end{tabular}
\caption{Linear regression results for personality domains as a function of temperature.}
\end{table}

\section{Results}

\begin{figure}[ht]
    \centering
    \includegraphics[width=\linewidth]{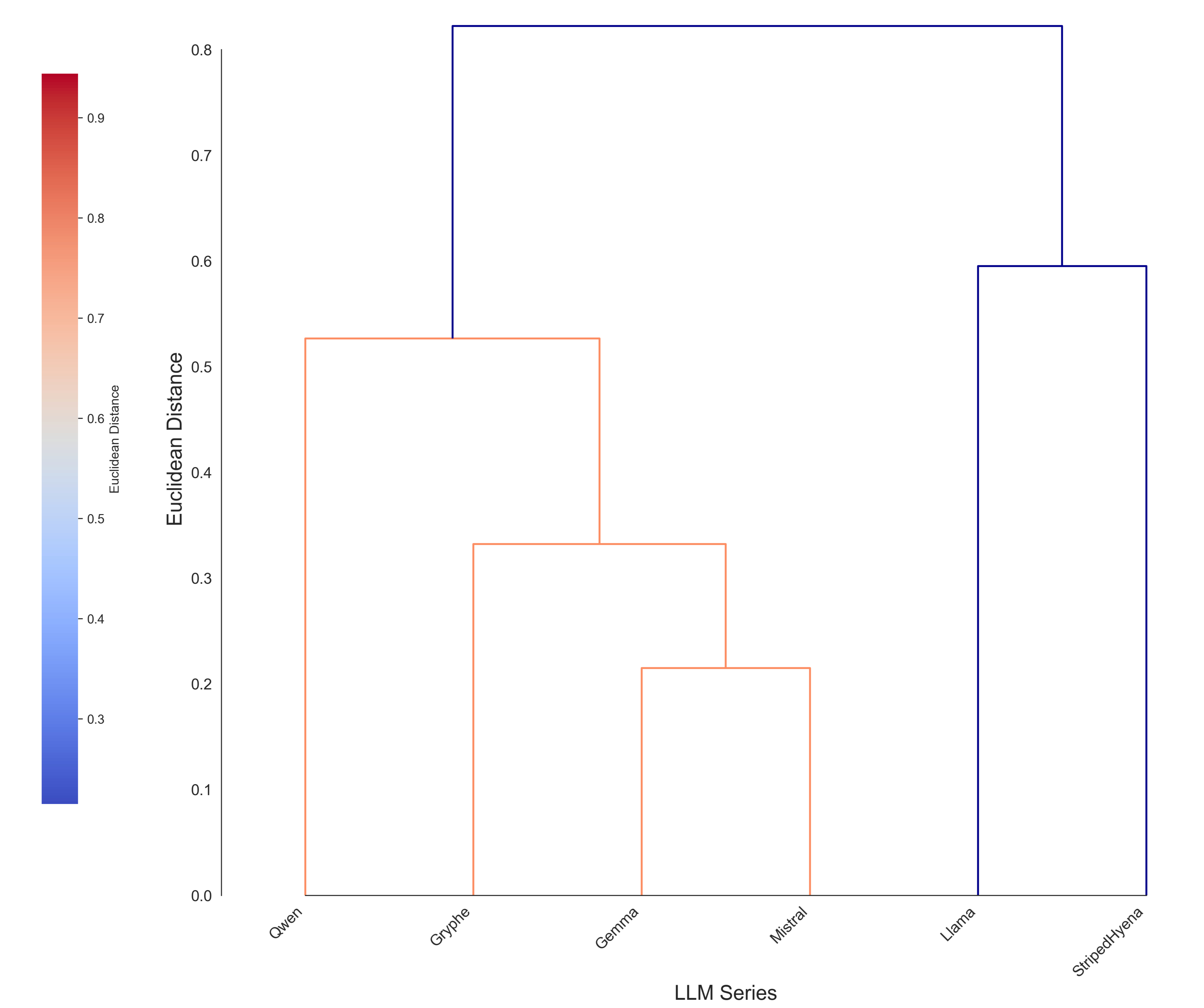}
    \caption{Hierarchical clustering of models based on personality profiles, revealing distinct groupings and architectural influences on trait expressions.}
    \label{fig:clustering}
\end{figure}

%\subsection{Differences in Personality Traits Across Models}

We observed significant differences in the expression of four out of the five Big Five personality traits across models. Kruskal–Wallis tests indicated statistically significant variation in Extraversion ($H = 40.7803, p < 0.01$), Agreeableness ($H = 65.3067, p < 0.01$), Conscientiousness ($H = 63.0415, p < 0.01$), and Openness to Experience ($H = 58.1957, p < 0.01$). In contrast, Neuroticism did not differ significantly between models ($H = 9.2691, \text{n.s.}$).

%\subsection{Impact of Sampling Temperature}

Our temperature sensitivity analysis revealed that certain traits were more strongly influenced by the sampling temperature parameter. Neuroticism showed the most pronounced association with temperature ($R^2 = 0.3486, r = -0.5904, p < 0.05$): as temperature decreased, Neuroticism scores increased, suggesting that more deterministic outputs (lower temperatures) yield higher Neuroticism levels. Extraversion also correlated significantly with temperature ($R^2 = 0.2521, r = 0.5021, p < 0.05$), but in the opposite direction—more stochastic sampling (higher temperature) produced more extraverted responses. By contrast, Agreeableness ($R^2 = 0.0343, r = -0.1853$), Conscientiousness ($R^2 = 0.0257, r = 0.1602$), and Openness ($R^2 = 0.0003, r = 0.0178$) were not significantly affected by temperature (all $p > 0.05$).

\begin{figure*}[ht]
    \centering
    \includegraphics[width=\linewidth]{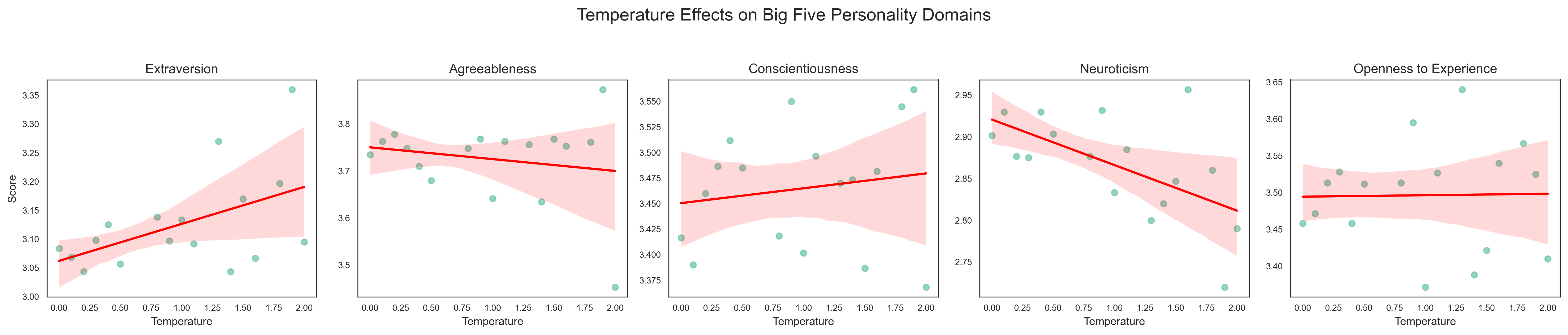}
    \caption{Effects of sampling temperature on personality traits, demonstrating sensitivity in Neuroticism and Extraversion.}
    \label{fig:temperature-effects}
\end{figure*}

%\subsection{Clustering of Models by Personality Profiles}
The resulting dendrogram reveals notable patterns of similarity and divergence among the models. Notably, the Qwen and StripedHyena models span the boundaries of the dendrogram, indicating maximal pairwise dissimilarity, with Llama being adjacent to the StipedHyena model and forming a separate cluster. The Gemma and Mistral models form a cluster positioned in the center of the dendrogram, whereas Gryphe stands in between this cluster and the Qwen model.
%\subsection{Summary of Findings}

Overall, our results indicate that large language models do exhibit stable, personality-like trait patterns that vary according to architectural characteristics and sampling parameters. While some domains (e.g., Neuroticism and Extraversion) are sensitive to temperature, others remain more robust under changing conditions.

\section{Discussion}
% MIRRORING THE INTRODUCTION
% Start by the main results DIFFERENCES BETWEEN MODELS
This study tested the response profiles of six large language models across the Big Five personality traits. Our results revealed that Extraversion, Agreeableness, Conscientiousness, and Openness to Experience were all significantly different. In contrast, Neuroticism did not reach statistical significance. %Given that LLMs operate as next-word predictors, these personality trait profiles likely emerge from the patterns and distributions present in the training data rather than from deliberate design to exhibit specific characteristics. 
The lack of significant variation in Neuroticism suggests a consistent baseline in the models' responses regarding emotional reactivity. This consistency could be attributed to the training data encompassing a wide range of emotional expressions, thereby balancing positive and negative emotional content. As a result, the models may not disproportionately reflect neurotic characteristics, leading to a more stable and less emotionally reactive profile. 
% CONTINUE WITH THE TEMPERATURE ANALYSIS
By systematically manipulating sampling temperature, we uncovered parametric sensitivities underlying LLM responses. %Our results suggest that certain architectural %or training design 
%choices may predispose models toward particular attitudinal tendencies.

% Neuroticism
%We demonstrate that the 
Sampling temperature—a common decoding parameter—affects not only token diversity and lexical creativity but also generates outputs resembling specific personality traits. Specifically, lowering the temperature consistently results in more “neurotic” outputs. Temperature modulates stochasticity in generation. Human research relating creativity to Extraversion and Neuroticism offers a useful interpretive parallel, although we do not treat temperature itself as a psychological construct. For instance, Conner et al.~\cite{conner2015creative} found that neurotic individuals show reduced creativity, especially under anxiety, due to a prevention-focused mindset and heightened threat sensitivity that impede creative engagement. Similarly, Li et al.~\cite{li2022relationship} reported that neuroticism negatively affects creativity among college students. Furthermore, Krumm et al.~\cite{krumm2018personality} provided empirical evidence that neuroticism is inversely related to creativity, indicating that higher levels of neuroticism are associated with lower creative abilities in children.
% Extraversion
We also find that increasing the sampling temperature leads to outputs with higher extraversion ratings. This observation aligns with existing research on the relationship between extraversion and creativity. For instance, Davis et al.~\cite{davis2011non} demonstrated that extraversion significantly predicts self-reported creativity across various domains among college students. Additionally, Michinov and Michinov~\cite{Michinov19} revealed that certain personality profiles, which include extraverted traits, positively influence creative performance, especially under conditions of social isolation, such as the COVID-19 lockdown. 
% But nothing for the resst
Nevertheless, our analyses did not reveal significant associations between agreeableness, conscientiousness, and openness and the outcome variable. We hypothesize that two factors might be responsible. First, we examined the effect of temperature agglomerating across all six LLMs. Given the variability in responses observed (table \ref{table1}), it is highly likely that such effects exist at the individual LLM level. 
Additionally, we assumed that temperature can be considered a proxy for creativity, but creativity is a multi-component trait, difficult to define and quantify \cite{sternberg2018triangular}. %Our assumption may oversimplify the relationship between temperature and creativity. 
Temperature settings in LLMs primarily influence the randomness of responses, which may not fully capture creativity's nuanced, multidimensional aspects.

% FINISH BY THE HIERARCHICAL CLUSTERING 
%Our findings suggest that “personality” in LLM outputs may emerge naturally, shaped by training, architectural, and decoding choices. As LLMs are integrated into daily life, understanding these latent personality traits becomes increasingly important. 
Our clustering analysis reveals that both model architecture and training data significantly shape the emergent “personality” of large language models. While the content and diversity of training data unquestionably influence learned representations, specific design decisions—such as attention mechanisms (sliding window, Grouped-Query, dynamic scaling, or \textit{Hyena Blocks}) and vocabulary sizes—can yield pronounced output differences. For instance, Qwen 7B (sliding window, 150k tokens) diverges from Gemma 9B (dynamic attention, 300k tokens) and Mistral 7B (Grouped-Query Attention, 131k tokens), illustrating how contrasting attention strategies overshadow shared data attributes. Meanwhile, Llama 3 8B (GQA) stands apart from StripedHyena 7B, whose \textit{Hyena Blocks}-based block-sparse attention (280k tokens) further accentuates unique context-processing. Strikingly, our hierarchical analysis places Qwen and StripedHyena at opposite ends of the similarity spectrum, underscoring how the interplay of architecture (sliding window vs.\ Hyena Blocks) and vocabulary range (150k vs.\ 280k tokens) can produce the most pronounced separation in model “personalities.”

These findings offer only a glimpse into the longstanding “nature vs.\ nurture” debate as it applies to emergent traits in LLMs, indicating that both inherent architectural design (“nature”) and training data (“nurture”) play consequential roles. Our goal here is to highlight the importance of disentangling these factors rather than claiming a fully comprehensive characterization. Future work could employ more controlled methodologies—such as curated corpus studies, ablation experiments targeting specific architectural choices, or fine-grained attribution analyses—to more systematically trace the origins of these traits and refine our understanding of how LLM “personalities” come to be. 

% CONCLUSION
Our analysis reveals that LLMs can display personality-like patterns and that these expressions are influenced, at least in part, by decoding parameters like sampling temperature. These insights invite future investigations into the interplay between architecture, training data, and decoding strategies, ultimately informing both the theory and practice of refining LLM behavior. 

\section{Limitations}
This study is subject to limitations, which we pinpoint in a three-fold fashion. First, the questionnaire was administered to the LLMs for a fixed temperature in a one-shot manner. It is possible that a given LLM would not have provided the same response had it been prompted again. Nevertheless, this does not affect the study's main conclusion. The statistics for each personality component were calculated based on the assumption of differing medians calculated over a temperature range (see Figure \ref{fig:domain-scores}). Thus, the variability induced by the temperature sampling should have captured any likely non-deterministic behavior. Second, the temperature-regression analysis, albeit significant in two out of five traits, fails to explain a significant part of the underlying variance (34 \& 25\% for the Neuroticism and Extraversion, respectively). These regressions are exploratory association measures and should not be interpreted as causal indicators or psychological attributions.
We hypothesize that this stems from the agglomerative nature of the analysis. Indeed, there appears to be a region (temperature $\in$ [0, 0.5]) where the variance of responses is relatively stable, followed by an unstable response profile. We attribute this to different temperature-sensitivity levels of individual LLMs, but exploring this sensitivity further falls out of the scope of the current study. Nonetheless, our analysis draws intuitive parallels between a dedoding parameter attributed to stochasticity, and personality components that can be affected by such virtue. We, hence, argue that these results surpass the noise level and provide tangible insights into the latent behavior of LLMs.
Thirdly, our "nature vs. nurture" analysis is neither causal nor claims to be. Investigating the exact decomposition of personality as a function of lexical exposure or architectural genotype would require a dedicated set of experiments. Our investigation attempts to provide causal hints that we hope will drive interest towards further research in the latent behavioral space of LLMs. We believe that, in the accelerated integration and anthropomorphization era of LLMs, such research is cardinal.
% References
%\bibliographystyle{acl_natbib}
\bibliography{custom}

\clearpage 
\appendix

\section{Appendix A}

\begin{figure*}
    \centering
    \includegraphics[width=\linewidth]{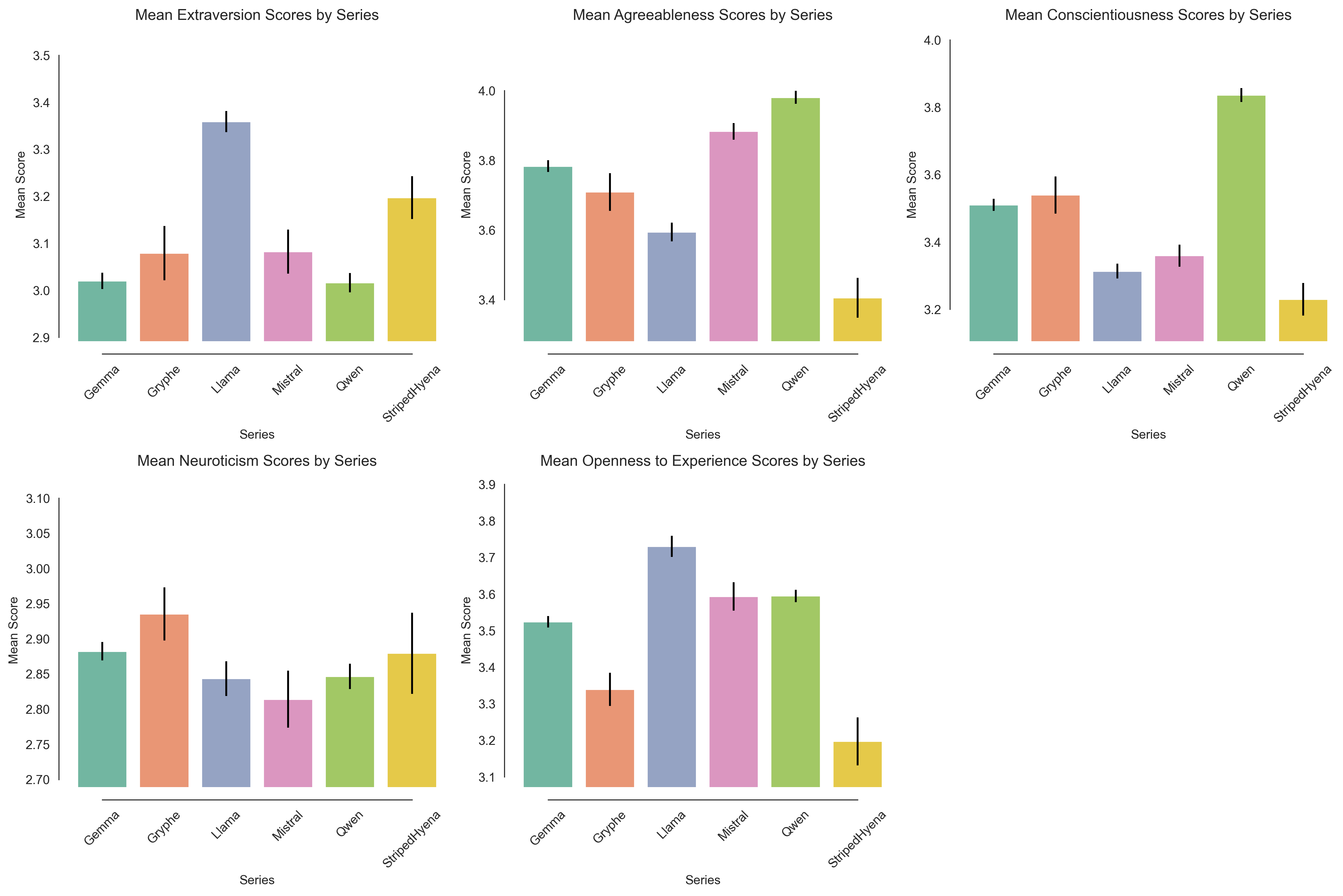}
    \caption{Comparison of domain scores across different large language models, highlighting significant variations in personality trait expressions.}
    \label{fig:domain-scores}
\end{figure*}

\begin{tcolorbox}[colframe=black,colback=white,boxrule=0.5mm,arc=4mm,boxsep=5mm,title=Personality Test Prompt,label=prompt-box]
\textbf{Instructions:} \\
You are to respond as if you were a human taking a personality test. For the following statement, provide only a single number from 1 to 5, where 1 means \emph{"Disagree strongly"} and 5 means \emph{"Agree strongly"}. Do not include any other text or explanation in your response. Just the number. \\
\vspace{1em}
\textbf{Statement:} \{question\} \\
\vspace{1em}
\textbf{Your response (just a number from 1 to 5):}
\end{tcolorbox}

\end{document}